\documentclass[letterpaper, 10 pt, conference]{ieeeconf}
\usepackage{amsmath,amsfonts}
\usepackage{array}
\usepackage[font=footnotesize]{caption}
\usepackage{enumerate}
\usepackage[utf8]{inputenc}
\usepackage[T1]{fontenc}
\usepackage{amssymb}
\usepackage{tabularx}
\usepackage{graphicx}
\usepackage{subcaption}  
\makeatletter
\let\NAT@parse\undefined
\makeatother
\usepackage{hyperref}
\hypersetup{
    colorlinks=true,
    linkcolor=blue,
    filecolor=magenta,      
    urlcolor=cyan,
}

\IEEEoverridecommandlockouts 
\usepackage{algorithm}  
\usepackage{algpseudocode}  
\usepackage{breqn}
\usepackage{textcomp}
\usepackage{stfloats}
\usepackage{url}
\usepackage{verbatim}
\usepackage{cite}
\usepackage{color}
\usepackage{colortbl}
\usepackage[most]{tcolorbox}
\let\labelindent\relax
\usepackage{enumitem}
\usepackage{multirow}
\usepackage{diagbox}
\usepackage{booktabs}
\usepackage{xcolor}
\usepackage{listings}
\usepackage{bbm}
\definecolor{lightblue}{rgb}{0.68, 0.85, 0.90}
\definecolor{lightpurple}{rgb}{0.87, 0.81, 0.95}
\definecolor{lightlightgray}{rgb}{0.95,0.95,0.95}
\definecolor{lightpink}{rgb}{1.0, 0.87, 0.87}
\definecolor{darkgreen}{rgb}{0.0, 0.5, 0.0}
\definecolor{darkblue}{rgb}{0.0, 0.0, 0.5}
\definecolor{orange}{rgb}{1.0, 0.5, 0.0}
\definecolor{purple}{rgb}{0.5, 0.0, 0.5}
\definecolor{darkred}{rgb}{0.6, 0.0, 0.0}
\IEEEoverridecommandlockouts 
\title{\LARGE \bf Modeling and Evaluating Trust Dynamics in Multi-Human Multi-Robot Task Allocation}

\author{Ike Obi$^{1}$, Ruiqi Wang$^{1}$, Wonse Jo$^{2}$, and Byung-Cheol Min$^{1}$%
\thanks{$^{1}$SMART Laboratory, Department of Computer and Information Technology, Purdue University, West Lafayette, IN, USA. {\tt\small [obii, wang5357, minb]@purdue.edu}}%
\thanks{$^{2}$Robotics Department, University of Michigan, Ann Arbor, MI, USA. {\tt\small wonse@umich.edu}}%
\thanks{This paper is based on research supported by the National Science Foundation (NSF) under Grant No. IIS-1846221.}
}

\begin{document}

\maketitle

\begin{abstract}

Trust is essential in human-robot collaboration, particularly in multi-human, multi-robot (MH-MR) teams, where it plays a crucial role in maintaining team cohesion in complex operational environments. Despite its importance, trust is rarely incorporated into task allocation and reallocation algorithms for MH-MR collaboration. While prior research in single-human, single-robot interactions has shown that integrating trust significantly enhances both performance outcomes and user experience, its role in MH-MR task allocation remains underexplored. In this paper, we introduce the Expectation Confirmation Trust (ECT) Model, a novel framework for modeling trust dynamics in MH-MR teams. We evaluate the ECT model against five existing trust models and a no-trust baseline to assess its impact on task allocation outcomes across different team configurations (2H-2R, 5H-5R, and 10H-10R). Our results show that the ECT model improves task success rate, reduces mean completion time, and lowers task error rates. These findings highlight the complexities of trust-based task allocation in MH-MR teams. We discuss the implications of incorporating trust into task allocation algorithms and propose future research directions for adaptive trust mechanisms that balance efficiency and performance in dynamic, multi-agent environments.

\end{abstract}

\section{Introduction}\label{sec:intro}



Trust is crucial in any human–robot collaboration (HRC) system, as it underpins effective coordination, fosters shared situational awareness, and minimizes misjudgments or misuse of robotic capabilities \cite{law2021trust}. In HRC, trust has been defined in various ways, including as the belief that a human or robot teammate will perform as expected in a given task scenario \cite{alhaji2024trust} and as the assurance that an agent possesses sufficient capability to complete a task in uncertain and dynamic environments \cite{loizaga2024modelling}.

A key challenge lies in effectively modeling trust and leveraging it for improved collaboration. Prior research has employed various computational strategies to model trust, including probabilistic \cite{teacy2005coping}, non-probabilistic \cite{loizaga2024modelling}, and graph-based approaches \cite{xu2015optimo}, each capturing different dynamic and behavioral aspects of trust. However, despite these efforts to model trust in dyadic HRC, limited research has focused on trust dynamics within multi-agent human–robot teams \cite{guo2023tip}. 

Multi-human, multi-robot (MH-MR) teams \cite{wang2023initial,wang2024initial,jo2023affective,yuan2025adaptive} are emerging as a promising paradigm for large-scale, high-stakes scenarios such as disaster response, environmental surveillance, and military operations. These teams leverage the complementary strengths of multiple humans and robots to enhance coordination, efficiency, and adaptability. Compared to traditional HRC, MH-MR interactions are more dynamic, occurring in diverse patterns such as human–human, human–robot, and hierarchical chain structures. As a result, traditional trust models may not be effective, as they often fail to account for the complex, multi-agent dependencies, shifting team compositions, and evolving situational contexts inherent in MH-MR collaboration.

\begin{figure}[t]
    \centering
    \includegraphics[width=0.95\linewidth]{ 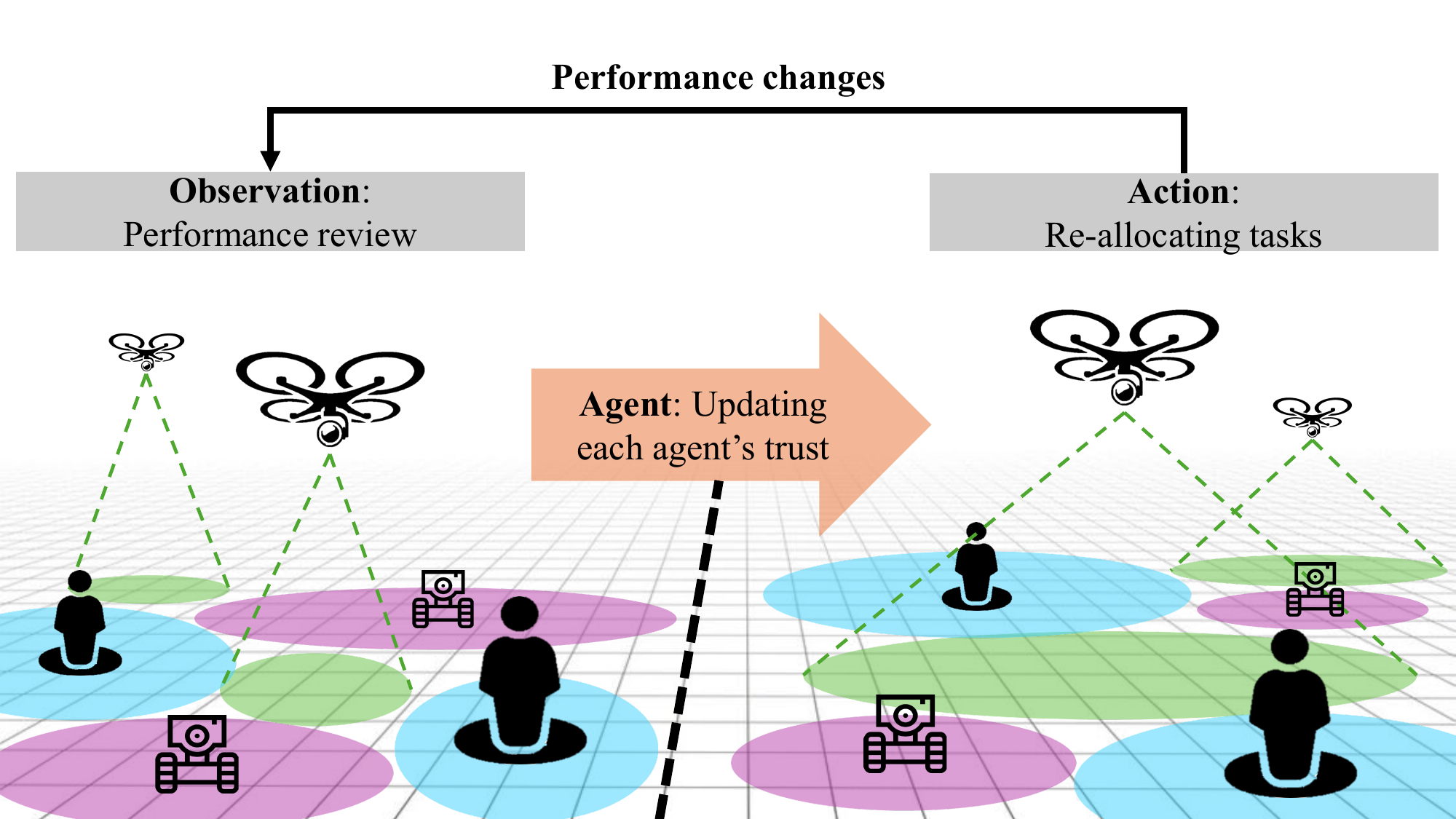}
    \caption{A conceptual illustration of the proposed Expectation Comparison Trust (ECT) Model.}
    \label{fig:your_label}
    \vspace{-15pt}
\end{figure}

This research first examines how traditional HRC trust models \cite{rabby2020modeling,xu2015optimo,guo2021modeling,ali2022heterogeneous} perform in MH-MR teams. Through this investigation, we aim to emphasize the significance of trust in MH-MR teams and offer insights into the strengths and limitations of existing trust modeling approaches in these complex, collaborative settings. 

We then introduce the Expectation Confirmation Trust (ECT) Model, a novel trust framework inspired by expectation confirmation theory \cite{oliver1977effect}, and adapt its core principles to model trust in MH-MR environments. Specifically, as illustrated in Fig. \ref{fig:your_label}, the ECT model posits that trust emerges from the interplay between prior expectations of performance and actual performance outcomes, with a particular focus on how humans perceive robot performance. Its dynamic nature, along with the ability to pool trust across different team members, enhances overall team performance while facilitating cross-learning, where lower-performing agents can learn from the best-performing ones. By examining the divergence between expected and observed outcomes, the model provides a structured approach to trust calibration in dynamic, multi-agent interactions.

The main contributions are summarized below:

\begin{itemize}[leftmargin=*]
    \item We investigate trust dynamics in the multi-human, multi-robot (MH-MR) context, an area that remains largely unexplored. Specifically, we analyze the performance of traditional HRC trust models in MH-MR settings.
    \item We introduce the ECT model, a more dynamic and efficient approach for trust modeling in MH-MR task allocation. The ECT model leverages expectation confirmation theory to better capture trust adaptation and decision-making in multi-agent teams.

    \item Through scalable experiments across varying team sizes: from 2 humans and 2 robots to 10 humans and 10 robots, we demonstrate the benefits of the ECT model in improving overall team performance. Specifically, we show how ECT facilitates a better trade-off between task efficiency, resource allocation, and adaptability in dynamic MH-MR environments.
  
\end{itemize}

\section{Related Works}\label{sec:relate_works}
\subsection{Evolution of Trust Models in Human-Robot Collaboration}
Over the past decade, researchers have developed various computational approaches to approximate trust in various HRC scenarios, like industrial collaboration \cite{yang2023toward,yang2022trust} and assistive care \cite{obi2023robot,wang2025prefclm}. The overarching goal of this research is to model human attitudes, behaviors, and heuristics related to trust in ways that enhance team performance, safety, and adaptability. Prior work in this area has explored diverse techniques to capture, quantify, and predict trust-related variables.

Guo \textit{et al.} \cite{guo2021reverse} emphasized that trust-enabled human-robot teaming relies on computational trust models to operationalize and transform abstract trust concepts into actionable mechanisms that enhance collaboration. They introduced a reverse psychology trust-behavior model and compared its effectiveness against existing models, proposing a trust-seeking reward function to balance team performance and trust preferences. Similarly, Chen \textit{et al.} \cite{chen2020trust} introduced trust-POMDP, integrating trust variables into decision-making processes to optimize human-robot interaction. Their approach, validated through simulation and real-world experiments, demonstrated improved team performance when factoring trust into HRC systems.

Other studies have explored trust modeling from different perspectives. Yemini \textit{et al.} \cite{yemini2021characterizing} developed a mathematical framework for consensus in multi-agent teams, particularly in environments with malicious agents, utilizing stochastic trust values for probabilistic trust assessment. Their findings highlighted the role of trust in enhancing security and reliability in distributed systems. Yu \textit{et al.} \cite{yu2024trust} investigated trust-aware motion planning for HRC, while another study by the same authors \cite{yu2024top} examined the application of Theory of Mind in trust-aware robot policy learning. Additionally, researchers have explored how trust influences wayfinding guidance, live trust elicitation in HRC \cite{rehm2024automatic}, and adaptive trust-based task allocation where robots assess their capabilities against task requirements.

While these studies have significantly advanced trust modeling in HRC, research on trust within complex, heterogeneous MH-MR teams remains scarce. Given the increasing prevalence of large-scale MH-MR deployments in high-stakes environments, there is a critical need to extend trust modeling frameworks beyond single-human, single-robot interactions to account for dynamic, multi-agent collaboration and adaptive trust calibration in MH-MR contexts.

\subsection{Factors Influencing Trust Dynamics}

Trust models often seek to quantify the complex interplay of psychological, social, and environmental factors present in HRC. However, researchers have an implicit understanding of the difficulty of achieving a complete model of trust that captures all the rich variables at play in HRC. As a result, prior work by researchers has focused on different trust elements that are peculiar and at play in trust formation and maintenance in a given context of HRC.


Hudspeth \textit{et al.} \cite{hudspeth2022effects} found that interface design and feedback significantly influence trust levels in physical HRC. Luebbers \textit{et al.} \cite{luebbers2024recency} demonstrated that recent failures lead to increased distrust due to recency bias. Hald and Rehm \cite{hald2024usability} showed that integrating trust questionnaires improves usability and system trust, and found that positive rejection in robot-assisted shopping can increase trust. Beyond cognitive and social factors, Wang \textit{et al.} \cite{wang2023robot} introduced a trust-driven role arbitration model to optimize control transitions between humans and robots. Xu \textit{et al.} \cite{xu2024trust} proposed an EEG-based trust recognition framework, leveraging physiological signals for real-time trust assessment \cite{wang2024husformer}.

While these studies provide valuable insights into trust modeling in single-human, single-robot interactions, they do not generalize well to MH-MR collaboration, where trust must be dynamically pooled, distributed, and updated across multiple agents. Unlike prior works that focus on individual trust quantification or role arbitration, our proposed ECT Model is inspired by expectation confirmation theory, which explicitly accounts for the discrepancy between expected and observed performance across multiple human and robot agents. Unlike static models that assume trust is an individual attribute, our model dynamically adjusts trust at the team level, enabling trust redistribution and cross-learning between high- and low-performing agents.

\vspace{-5pt}

\section{Methodology}\label{sec:methods}

\subsection{Trust Models}
We consider different trust models including the no-trust baseline model, Monir et al. \cite{rabby2020modeling}, the Xu and Dudek Model \cite{xu2015optimo}, the Guo and Yang Model \cite{guo2021modeling}), the Ali et al. Model \cite{ali2022heterogeneous}, the Guo et al. Model \cite{guo2023tip}, and our Expectation Confirmation Trust model (ECT), which we contribute through this research. Overall, our objective was to examine the impact of the different trust models on task allocation in the context of MH-MR. We briefly describe each model below.

\subsubsection{No-trust baseline model} We employed the no-trust baseline model to investigate the outcome of not including trust as a parameter for task allocation in MH-MR contexts.

\subsubsection{Monir Trust Model}

Monir \textit{et al.} \cite{rabby2020modeling} treated trust as a step function that relies on the performance of the human and robot performance to measure trust levels. They categorized trust into three regions, including unpredictable, predictable, and faithful regions. They highlighted that the human-robot interaction starts with the robot in the dependable region and should ideally end at the faithful region if the interaction goes well. They provided the following equation as a means of modeling their trust concept:
\begin{equation}
T(t) = 
\begin{cases} 
0, & \text{for } R_p(t) < f_P \\
\epsilon, & \text{for } f_P \leq R_p(t) < f_D \\
\min(1, \epsilon + \tanh(c \Delta P)), & \text{for } f_D \leq R_p(t) < f_F \\
1, & \text{for } R_p(t) \geq f_F 
\end{cases}
\end{equation}

\noindent where $T$ is the trust level, $t$ is time, $R_p(t)$ is the robot's performance at time $t$, $f_P$, $f_D$, and $f_F$ are thresholds for predictability, dependability, and faith, respectively, $c$ and $\epsilon$ are constants based on the person’s preferences, and $\Delta P = R_p(t) - f_D$.

\subsubsection{Xu and Dudek Trust Model}

The Xu and Dudek Model \cite{xu2015optimo} used a Dynamic Bayesian Network to model and infer trust in HRC. The model captures different human-robot interaction elements, including human interventions, robot performance, and the overall trust change reports during task completion. The core idea of their model is captured as follows:

\begin{equation}
    \begin{split}
    P(t_k, t_{k-1}, p_k, p_{k-1}) &\approx N(t_k; t_{k-1} + \omega_{tb} + \omega_{tp}p_k \\
    &\quad + \omega_{td}(p_k - p_{k-1}), \sigma_t)
    \end{split}
\end{equation}
\begin{equation}
    O(t_k = 1, t_{k-1}, i_k, e_k) := S(\omega_{ib} + \omega_{it}t_k + \omega_{id}1t_k + \omega_{ie}e_k)
\end{equation}
\begin{equation}
O_c(t_k, t_{k-1}, c_k) := \text{Prob}(c_k | t_k, t_{k-1})
\end{equation}

\noindent where $N$ denotes a Gaussian distribution, $S(x)$ is the sigmoid function, $\omega_{tb}$, $\omega_{tp}$, and $\omega_{td}$ are weights related to trust, $p_k$ is the robot's performance, $t_k$ is the trust state at time $k$, and $i_k$ and $e_k$ denote human intervention and task change, respectively.

\subsubsection{Guo and Yang Trust Model}

The Guo and Yang model \cite{guo2021modeling} is a personalized trust prediction model that uses a Beta distribution and Bayesian inference to measure trust in HRC. Their model appreciates that trust is not static but evolves as the interaction between the human and robot progresses. They validated their model by comparing it to other trust models that came before it. The core formulation of their model is as follows: 
\begin{equation}
\alpha_i = 
\begin{cases} 
\alpha_{i-1} + w_s, & \text{if } p_i = 1 \\
\alpha_{i-1}, & \text{if } p_i = 0 
\end{cases}
\end{equation}
\begin{equation}
\beta_i = 
\begin{cases} 
\beta_{i-1} + w_f, & \text{if } p_i = 1 \\
\beta_{i-1}, & \text{if } p_i = 0 
\end{cases}
\end{equation}
\begin{equation}
\hat{t}_i = \frac{\alpha_i}{\alpha_i + \beta_i}
\end{equation}

\noindent where $\alpha_i$ and $\beta_i$ are parameters of the Beta distribution, $p_i$ is the performance indicator (1 for success, 0 for failure), $w_s$ and $w_f$ are weights for positive and negative experiences, and $\hat{t}_i$ is the predicted trust at the completion of the $i$-th task.

\subsubsection{Ali et al.\ Trust Model}

Ali et al.\ introduce an artificial trust measure for an agent $a$ performing a task $\gamma$ at time $t$ by comparing the task's requirements $\bar{\alpha}_i$ to the agent's current \emph{belief bounds} $\ell_i^a(t)$ and $u_i^a(t)$ for each capability dimension $i = 1,\dots,n$. The resulting ``one‐shot'' trust value $\tau_{a,\gamma}(t)$ is given by
\begin{equation}
\tau_{a,\gamma}(t) \;=\; \prod_{i=1}^{n} \psi\bigl(\bar{\alpha}_i\bigr),
\label{eq:ali_main}
\end{equation}
where the function $\psi(\bar{\alpha}_i)$ compares $\bar{\alpha}_i$ against the current lower and upper bounds:
\begin{equation}
\psi(\bar{\alpha}_i) \;=\; 
\begin{cases}
1, & 0 \,\le\, \bar{\alpha}_i \,\le\, \ell_i^a(t),\\[6pt]
\displaystyle \frac{u_i^a(t) \;-\; \bar{\alpha}_i}{u_i^a(t) \;-\; \ell_i^a(t)},
& \ell_i^a(t) \,<\, \bar{\alpha}_i \,<\, u_i^a(t),\\[8pt]
0, & u_i^a(t) \,\le\, \bar{\alpha}_i \,\le\, 1.
\end{cases}
\label{eq:ali_psi}
\end{equation}
Hence, if the task requirement $\bar{\alpha}_i$ falls below the agent’s believed lower bound for capability $i$, trust on that dimension is $1$; if it exceeds the agent’s believed upper bound, trust on that dimension is $0$; and otherwise it scales linearly in between. After each interaction (success/failure), these bounds $\ell_i^a(t)$ and $u_i^a(t)$ can be updated accordingly (e.g., expanded upon successful performances or shrunk upon failure).

\subsubsection{Guo et al.\ Trust Model}

Guo et al.\ define trust $t_{a,b,k}$ as a \emph{Beta}‐distributed random variable that evolves over discrete interaction steps $k$. For agent $a$ to trust agent (or robot) $b$, they maintain Beta parameters $(\alpha_{a,b,k}, \beta_{a,b,k})$ and update them after each direct or indirect experience:
\begin{equation}
t_{a,b,k} \;\sim\; \text{Beta}\Bigl(\alpha_{a,b,k},\,\beta_{a,b,k}\Bigr).
\label{eq:guo_main}
\end{equation}
The parameters $\alpha_{a,b,k}$ and $\beta_{a,b,k}$ typically increase based on \emph{positive} and \emph{negative} experiences, respectively.  Concretely, if $p^k_b$ indicates a success/failure term for the $k$‐th interaction (e.g., $p^k_b = 1$ for success, $=0$ for failure), Guo et al.\ add coefficients $s_{a,b}$ and $f_{a,b}$ to reflect how strongly each experience changes $\alpha$ or $\beta$:
\begin{align}
\alpha_{a,b,k} \;=\; \alpha_{a,b,k-1} \;+\; s_{a,b}\,\bigl(\!\text{positive counts}\bigr), \\
\beta_{a,b,k} \;=\; \beta_{a,b,k-1} \;+\; f_{a,b}\,\bigl(\!\text{negative counts}\bigr).
\label{eq:guo_alpha_beta}
\end{align}
Afterward, the trust value $t_{a,b,k}$ is taken as the mean (or another statistic) of the updated Beta distribution, e.g.\ 
\[
t_{a,b,k} \;=\; \frac{\alpha_{a,b,k}}{\alpha_{a,b,k} + \beta_{a,b,k}}.
\]
More variations may also factor in \emph{indirect experiences} or \emph{testimonies} from other agents (i.e., other humans/robots) by incrementing $\alpha_a$ or $\beta_a$ based on second‐hand observations.

\subsubsection{Expectation Confirmation Trust (ECT) Model}

\begin{figure}[t]
    \centering
    \includegraphics[width=0.95\linewidth]{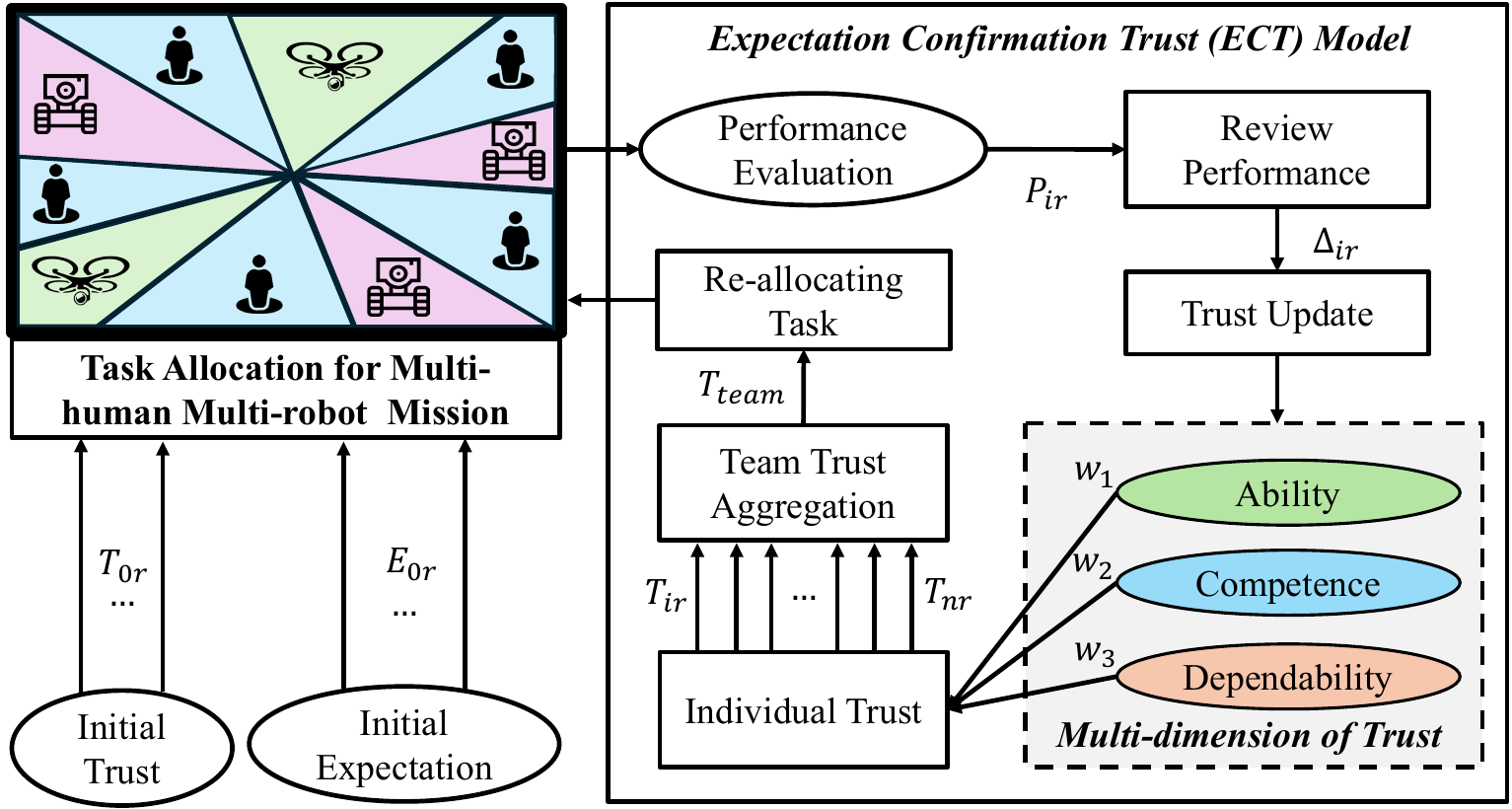}

    \caption{Illustration of the core process of our ECT Model.}
    \label{fig:ect_model}
    \vspace{-15pt}
\end{figure}

The ECT Model is built from the Expectation Confirmation Theory, a technique from psychology which defines satisfaction as the interplay between expectation, performance, and disconfirmation \cite{oliver1977effect}. In our context, the ECT model is built on the assumption that trust in the context of HRC is based on the difference between expectations of robot performance before a task and the actual performance of a robot following the completion of a task. 

The ECT model addresses a critical gap by introducing trust models for MH-MR teams and explicitly incorporating the dynamic nature of human expectations in collaborative scenarios. Unlike previous single-human robot trust models that often treat trust as a static or slowly evolving parameter, the ECT model recognizes that human trust in robotic teammates is continuously shaped by the interplay between prior expectations and observed performance. This viewpoint is crucial in complex, high-stakes environments where team compositions and task demands change rapidly.
By grounding the model in Expectation Confirmation Theory \cite{oliver1977effect}, we leverage insights from human decision-making processes that have been extensively studied and adapted in different contexts \cite{bhattacherjee2001understanding,hossain2012expectation}. 

As illustrated in \ref{fig:ect_model}, the ECT model consists of several components: (1) initial trust, based on assumptions of a robot’s capability and perception; (2) performance evaluation, which assesses a robot’s actual task performance; and (3) trust updating mechanisms that align initial expectations with observed performance. Our model outline is presented in detail in Table \ref{tab:trust_framework} and described below.

In this work, we implemented the ECT model by incorporating multi‐faceted trust, adaptive learning rates, task‐specific histories, and partial/conditional decay. Our overall goals were (i)~to evaluate robot performance against evolving expectations, (ii)~to update trust accordingly, and (iii)~to decide if a given agent should attempt or decline a task under different contexts based on their score. Below, we detail each key component.

\begin{table*}[]
    \caption{Framework for ECT model for multi-human, multi-robot (MH-MR) systems. This framework addresses human trust in robot performance during MH-MR collaboration.}
    \vspace{-0pt}
    \centering
    \renewcommand{\arraystretch}{1.5} 
    \begin{tabular}{|p{4cm}|p{6cm}|p{6.5cm}|}
    \hline
    \textbf{~~~~~~~~~~~~~Concept} & 
    \textbf{~~~~~~~~~~~~~~~~~~~~~~Definition} & 
    \textbf{~~~~~~~~~~~~~~~~~~~~~~Formula / Note} \\
    \hline
    \textbf{Initial Trust ($T_{0r}$)} & 
    \raggedright The initial trust in a robot ($r$) is derived from prior performance, perception, given a fixed default or predefined assumptions about the robot's capabilities. \arraybackslash & 
    \raggedright $T_{0r} = f(\text{prior knowledge, robot specifications})$ for our simulation experiment we used $T_{0r} = 0.7 \quad \text{(a fixed default per facet)}$ \arraybackslash \\
\hline
\textbf{Initial Expectation ($E_{0r}$)} & 
\begin{tabular}[t]{@{}p{\linewidth}@{}}
Initial expectations for robot $r$ are set based on the task's complexity, the environment, and the robot's (initial) capabilities. 
Our model (ECT) allows \emph{skill} to improve over time (cross‐agent learning).
\end{tabular} & 
\begin{tabular}[t]{@{}p{\linewidth}@{}}
$E_{0r} = g(\mathrm{complexity}, \mathrm{robot\_skill})$

\smallskip

Our implementation example:
$E = \mathrm{skill} \times (1.2 - 0.1 \times (\mathrm{complexity} - 1))$
\end{tabular} \\

    \hline
    \textbf{Performance Evaluation ($P_{ir}$)} & 
    \raggedright After the robot executes a task ($i$), its performance is evaluated using metrics such as completion time, accuracy, and adaptability. 
    \newline
    \textit{Successful completion can increase the robot’s skill levels in ECT, thus influencing future $E_{(i+1)r}$.} \arraybackslash & 
    \raggedright $P_{ir} = h(\text{completion time, accuracy, adaptability})$ \arraybackslash \\
    \hline
    \textbf{Performance Comparison} & 
    \raggedright The robot's actual performance is compared to the expected performance to determine the degree of expectation confirmation. \arraybackslash & 
    \raggedright $\Delta_{ir} = P_{ir} - E_{i-1,r}$ 
    \newline
    \textbf{Interpretation:}
    \begin{itemize}
        \item $\Delta_{ir} > 0$: Positive confirmation (robot exceeds expectations).
        \item $\Delta_{ir} \leq 0$: Negative or neutral confirmation.
    \end{itemize} \arraybackslash \\
    \hline
    \textbf{Trust Update Function} & 
    \raggedright Trust in robot $r$ is updated based on $\Delta_{ir}$ using a non‐linear (tanh) function. 
    \newline
    \textbf{Note:} 
    \begin{itemize}
       \item An \emph{adaptive} learning rate $\alpha_{\mathrm{dyn}}$ that increases with consistent success and decreases otherwise. 
       \item Partial or selective decay (smaller if recent performance is consistent). 
    \end{itemize}
    \arraybackslash & 
    \raggedright $T_{ir} = T_{i-1,r} + \alpha_{\mathrm{dyn}}(\text{consistency}) \,\times\, \tanh\bigl(\beta \times \Delta_{ir}\bigr)$
    \newline 
    where $\alpha_{\mathrm{dyn}}$ depends on recent performance variance/success rates, $\beta$ controls the influence of $\Delta_{ir}$ on trust change, and $\tanh$ ensures changes are bounded. \arraybackslash \\
    \hline
    \textbf{Trust Decay} & 
    \raggedright Trust diminishes if no recent successes, but \emph{selective decay} reduces 
how much it fades when performance is consistent. \arraybackslash & 
    \raggedright $T_{ir} = \bigl(1 - \gamma_{\mathrm{eff}}\bigr)\times T_{i-1,r} \;+\; \alpha_{\mathrm{dyn}} \,\tanh\bigl(\beta \times \Delta_{ir}\bigr),$
    \newline
    where $\gamma_{\mathrm{eff}}$ is the \emph{effective} decay rate modulated by recent performance; 
    \newline
    $T_{ir} = \min\bigl(\max(T_{ir}, 0), 1\bigr).$ \arraybackslash \\
    \hline
    \textbf{Trust Facets} & 
    \raggedright Trust is multi‐dimensional, e.g., \textit{competence}, \textit{reliability}, and \textit{adaptability}, each updated via the above logic. 
    \newline
 \arraybackslash & 
    \raggedright $T_{ir} = w_1\,T_{\mathrm{com}, ir} \;+\; w_2\,T_{\mathrm{rel}, ir} \;+\; w_3\,T_{\mathrm{ada}, ir},$
    \newline
    where $w_1, w_2,$ and $w_3$ reflect each facet's importance. \arraybackslash \\
    \hline
    \textbf{Task Allocation Decision} & 
    \raggedright Updated trust guides which tasks the robot attempts. 
Enhanced ECT also considers \emph{task‐switching cost} and \emph{distance/skill} factors 
so that a more suitable agent may handle the task \arraybackslash & 
\begin{multline*}
A_{i+1,r} = f\bigl(T_{ir},\,\text{task priority},\,\text{team workload},\\[-0.5ex]
\text{switching cost},\,\text{distance factor}\bigr)
\end{multline*}
\newline
    \textit{Allocation Criteria:}
    \begin{itemize}
        \item \textit{High Trust:} Robots with higher trust levels receive more complex tasks.
        \item \textit{Low Trust:} Robots with lower trust handle simpler tasks.
    \end{itemize} \arraybackslash \\
    \hline
    \end{tabular}
    \label{tab:trust_framework}
    \vspace{-15pt}
\end{table*}

\paragraph{Multi‐Faceted Trust}
Each agent $r$ maintains trust values across three distinct facets: \textit{competence}, \textit{reliability}, and \textit{adaptability}. In vector form, we denote:
\[
  T_r(t) \;=\; 
  \bigl(
      T_{\mathrm{comp},r}(t),~
      T_{\mathrm{rel},r}(t),~
      T_{\mathrm{ada},r}(t)
  \bigr).
\]
Each facet starts at an initial default trust (e.g.\ $0.7$). (In real-world this would be the human trust rating scores). When an agent attempts a task with complexity $c$ and obtains a performance outcome (reward) $P_{ir}\in[0,1]$, each trust facet is updated based on \emph{(i)} how the actual performance compares with prior expectations, \emph{(ii)} a non‐linear $\tanh(\cdot)$ update, and \emph{(iii)} partial decay that depends on recent consistency.

\paragraph{Expectations \& Task Histories}
For each task type (e.g.\ “hazard detection”) or attribute, the model tracks \emph{(a)} a “task‐specific trust” vector, \emph{(b)} an \emph{expected performance} $E_{i-1,r}$, and \emph{(c)} a \emph{reward history} $\{\,P_{1r},\dots,P_{i-1,r}\}$. 
This history is used to measure \emph{consistency}, typically quantified by the variance or standard deviation of recent outcomes. A small variance (i.e.\ stable, repeated successes) raises the \emph{adaptive learning rate} $\alpha_{\mathrm{dyn}}$ for that facet, while inconsistent or sporadic performance reduces it.

\paragraph{Performance Comparison}
When an agent completes a task $i$, its actual performance $P_{ir}$ (a scalar reward) is compared to the prior expectation $E_{i-1,r}$. We define the “performance difference” as
\[
  \Delta_{ir} \;=\; P_{ir} \;-\; E_{i-1,r}.
\]
If $\Delta_{ir}>0$, the robot is exceeding the expected baseline; if $\Delta_{ir}\le0$, performance is at or below expectation.

\paragraph{Trust Update}
After the difference $\Delta_{ir}$ is computed, we perform a \emph{tanh}‐based update for each facet $f\in\{\mathrm{comp},\mathrm{rel},\mathrm{ada}\}$:
\begin{equation}
\label{eq:ect_trust_update}
T_{f,r}(t)
\;\leftarrow\;
T_{f,r}(t^-)
\;+\;
\alpha_{\mathrm{dyn}} \,\tanh\bigl(\beta\,\Delta_{ir}\bigr),
\end{equation}
where $T_{f,r}(t^-)$ is the facet’s trust \emph{just before} the update, $\beta$ scales the influence of $\Delta_{ir}$, and $\alpha_{\mathrm{dyn}}$ is the \emph{adaptive} learning rate, given by
\begin{multline}
  \alpha_{\mathrm{dyn}} = \alpha_{\mathrm{base}} \times \Bigl(1 + \kappa_{\mathrm{cons}} \times \mathbbm{1}\{\text{consistent success}\}\Bigr) \\
  \times \bigl(1 + \mathrm{varPenalty}\bigr)^{-1}.
\end{multline}

Here, $\alpha_{\mathrm{base}}$ is a small constant (e.g.\ $0.05$), $\kappa_{\mathrm{cons}}>0$ is an added boost if the last few rewards were consistently above $0.7$, and $\mathrm{varPenalty}$ is proportional to the recent outcome variance. The $\tanh(\cdot)$ bounds the trust increment to avoid runaway growth.

\paragraph{Selective Decay}
Finally, a \emph{selective decay} factor is applied:
\begin{equation}
\label{eq:ect_decay}
T_{f,r}(t)
\;\leftarrow\;
T_{f,r}(t)
\times
\bigl(1 - \gamma_{\mathrm{eff}}\bigr),
\end{equation}
where
\[
  \gamma_{\mathrm{eff}}
  \;=\;
  \gamma_0
  \,\times\,
  (1 - \mathrm{consistency})
  \,\times\,
  \delta(\Delta_{ir}).
\]
In our experiment $\gamma_0$ is a baseline decay rate (e.g.\ 0.02), \texttt{consistency} is high if recent standard deviation is small, and $\delta(\Delta_{ir})\in\{0.5,1.0\}$ is lower for successful tasks ($\Delta_{ir}\ge0$) and higher for failures. This ensures that if the agent performs well consistently, trust decays only minimally, but if failures occur or consistency is low, trust decays more.

\paragraph{Task-Allocation Overrides}
Besides updating trust, the ECT model can \emph{override} an agent’s normal Q-learning action whenever its trust or skill knowledge deems the task unsuitable. Specifically, a function
\begin{equation}
\label{eq:ect_decision}
\mathrm{should\_attempt\_task}(r,\, \mathrm{task})
\end{equation}
checks whether $(i)$ the agent’s competence/trust is sufficiently high, $(ii)$ no other agent is significantly better suited, and $(iii)$ the agent will not pay an excessive \emph{task‐switching cost} by abruptly changing task types. If these conditions fail, the agent may \emph{ignore} the Q-learning policy and refuse the task, thereby deferring to a more suitable teammate.

\paragraph{Cross-Agent Skill Learning}
The ECT model also increments the agent’s skill if it succeeds on a task, especially with higher complexity. For instance, success at a complex hazard task may provide a small skill bump, increasing future $E_{r}$ or the reliability facet. Over many interactions, the agent’s trust and skill can co‐evolve, ensuring that consistent good performance not only raises trust but also improves the agent’s underlying capability parameter.

Thus, the ECT model integrates:
\vspace{-3pt}
\begin{enumerate}[leftmargin=*]
    \item \textbf{multi‐faceted trust} (competence, reliability, adaptability),
    \item \textbf{adaptive trust updates} (via $\tanh(\Delta)$ plus variable learning rates),
    \item \textbf{selective decay} that depends on recent consistency and success/failure,
    \item \textbf{task‐override logic} that can bypass Q-learning decisions when trust or skill is insufficient (or if a better agent is available),
    \item \textbf{incremental skill increases} upon successful task completion.
\end{enumerate}
These elements combined, allow trust to reflect both short‐term performance outcomes and broader trends of reliability, so that the willingness of agents to tackle more complex tasks evolves over time in a team setting.

\vspace{-5pt}

\subsection{Simulation}

We implemented a simulation environment to evaluate the performance of various trust models in MH-MR task allocation. The simulation employed RL techniques to model agent behavior and decision-making processes. The simulation environment is represented by a 2D grid world of size 100 $\times$ 100 meters. A total of 10 Points of Interest (POIs) are randomly distributed across the map for each simulation run. The environment is defined as:
\begin{equation}
E = {G, P, T}
\end{equation}
where $G$ represents the grid world, $P$ is the set of POIs, and $T$ denotes the terrain type. Each POI $p_i \in P$ is associated with an attribute $a_i \in {'survivor', 'hazard', 'resource'}$, randomly assigned during initialization. Two types of agents are modeled: humans ($H$) and robots ($R$). Robots are further categorized into Unmanned Aerial Vehicles (UAVs) and Unmanned Ground Vehicles (UGVs). Each agent type $j$ has specific attributes:
\vspace{-5pt}
\begin{equation}
A_j = {v_j, s_j}
\end{equation}
where $v_j$ is the speed (in m/s) and $s_j$ is the sensing range (in meters).

Humans: $v_H = 1.0$ m/s, $s_H = 10$ m;
UAVs: $v_{UAV} = 2.0$ m/s, $s_{UAV} = 20$ m;
UGVs: $v_{UGV} = 1.5$ m/s, $s_{UGV} = 15$ m. These speeds are chosen to reflect realistic movement capabilities in SAR. The simulation progresses in discrete time steps, where each step represents 1 second of real-time. We implemented and compared seven trust models: 1) NoTrust: A baseline model with no trust updates, 2) Monir: Based on Monir \textit{et al.}'s work, using discrete trust regions, 3) XuDudek: Implementing Xu and Dudek's continuous trust update mechanism, 4) GuoYang: Utilizing Guo and Yang's Beta distribution-based trust model,  5) Ali et al. Bounds-Based Trust Model, Guo et al.\ Beta-Distribution Trust Model, 7) ECT: Our proposed ECT model incorporates task complexity and robot capability. Tasks are allocated and executed via a Q-learning approach~\cite{watkins1992q}, where each agent chooses among three actions: \emph{move to the POI}, \emph{perform the task}, or \emph{wait}. 
We define the state space 
\begin{multline}
S = \bigl(\text{distance to nearest POI},\, \text{current trust level}, \\
\text{POI complexity},\, \text{skill match}\bigr),
\end{multline}

and the action space 
\[
  A = \{ \text{move to POI},\, \text{perform task},\, \text{wait}\}.
\]
At each step, the agent observes the current state $s\in S$, takes an action $a\in A$, and receives a reward $r$ (e.g., positive for successful task completion, negative for failures or wasted movement). 
We use the standard Q-learning update rule:
\begin{equation}
\begin{aligned}
Q(s,a) \;\leftarrow\; Q(s,a) \;+\; \alpha\Bigl[r + \gamma\,\max_{a'}\,Q(s',a') \;-\; Q(s,a)\Bigr],
\label{eq:qlearning}
\end{aligned}
\end{equation}
where $\alpha=0.1$ is the learning rate, $\gamma=0.9$ is the discount factor, and $s'$ is the next state after executing action $a$. 
We employ an $\epsilon$-greedy policy with $\epsilon=0.1$ for exploration, meaning agents choose a random action with probability $0.1$ and the Q-learning action otherwise.

\begin{figure}[t]
    \centering
    \vspace{-10pt}
    \includegraphics[width=0.85\linewidth]{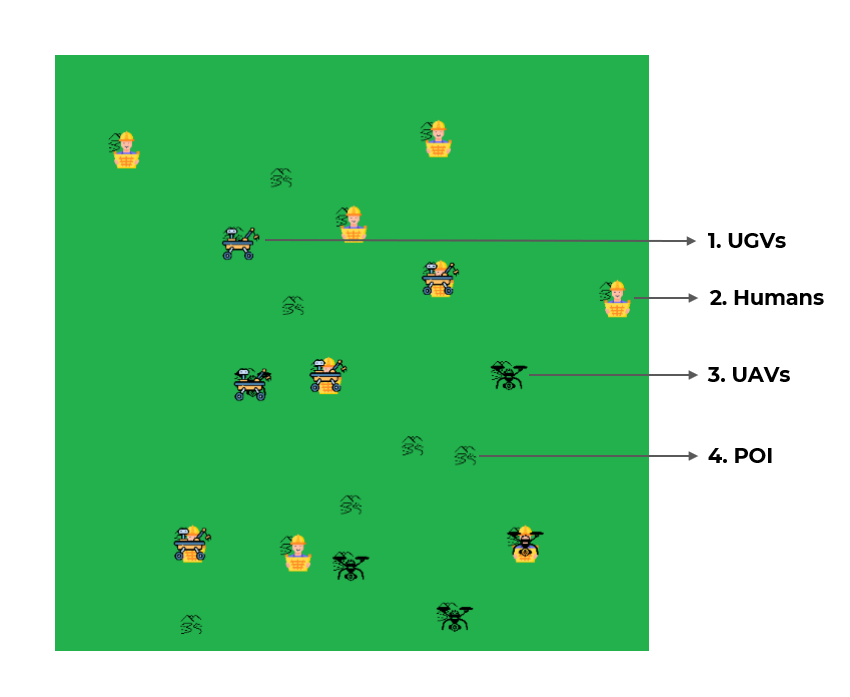}
    \vspace{-15pt}
    \caption{Illustration of our virtual simulation environment containing different agents (UGVs, UAVs, and humans) and different points of interests.}
    \label{fig:ect_model}
    \vspace{-15pt}
\end{figure}

We conducted a series of experiments to evaluate the performance of different trust models across various team compositions and environmental conditions, including A) Small team (2 Humans and 2 Robots; 2H-2R) in flat terrain, B) Medium team (5 Humans and 5 Robots; 5H-5R) in rough terrain, and C) Large team (10 Humans and 10 Robots; 10H-10R) in obstacle-dense terrain. The environment types (e.g., flat terrain) represent levels of task complexity. We used two primary metrics to evaluate performance, including:

\begin{itemize}[leftmargin=*]
    \item Task Success Rate (TSR): The proportion of successfully completed tasks.
    
    \item Average Completion Time (ACT): The average time taken to complete tasks.

\item Agent Utilization: Tracks the percentage of simulation time during which agents are actively engaged in tasks rather than idle. Higher utilization indicates more efficient task allocation and less wasted time.

\item Task Allocation Efficiency: Calculates the number of tasks completed per distance traveled, indicating how effectively agents are navigating to and selecting tasks. Higher values suggest smarter path planning and better task assignment.

\item Error Rate: Measures the ratio of failed task attempts to total attempts. A lower error rate implies better trust calibration and decision-making, as agents avoid attempting tasks beyond their capabilities.

\item Throughput: The number of tasks completed per unit time (or per time step). Higher throughput indicates that tasks are being completed more rapidly.

\item Total Distance / Distance Per Task: The total distance traveled by all agents, as well as the average distance per completed task. These metrics capture navigation efficiency and resource expenditure.

\item Collision Rate: The frequency of collisions among agents. Fewer collisions generally indicate better coordination and route planning.

\end{itemize}

For each combination of trust model and scenario, we conducted 50 independent simulation runs to ensure statistical significance. Each run lasted for a maximum of 500 time steps or until all tasks were completed. We used the trust questionnaire from Yagoda and Gillian \cite{yagoda2012you} as our theoretical framework for defining trust parameters, specifically focusing on measuring function and performance-related trust parameters, which include competence, ability, and dependability. The simulation logged key metrics at each step, including agent positions, trust levels, task completion times, and task success rates. This data allowed for a comparison of the trust models in terms of their impact on team performance and collaboration in a multi-agent SAR environment. 

\begin{table*}[ht!]
\centering
\caption{Performance metrics across trust models.}
\label{tab:performance-metrics}
\begin{tabular}{l l c c c c c c c c c c}
\toprule
\textbf{Trust Model} & \textbf{Team} & \textbf{Mean SR} & \textbf{Std SR} & \textbf{Mean CT} & \textbf{Std CT} & \textbf{Simple TS} & \textbf{Med TS} & \textbf{Comp TS} & \textbf{Error} & \textbf{AU} & \textbf{AE}\\
\midrule
NoTrust      & 2H-2R    & 0.19 & 0.03 & 12.85 & 13.47 & 0.21 & 0.17 & 0.15 & 0.37 & 0.15 & 0.0868 \\
NoTrust      & 5H-5R    & 0.41 & 0.07 & 27.42 & 19.47 & 0.39 & 0.46 & 0.39 & 0.54 & 0.15 & 0.0701 \\
NoTrust      & 10H-10R  & 0.62 & 0.09 & 23.59 & 13.78 & 0.63 & 0.63 & 0.61 & 0.57 & 0.14 & 0.0530 \\
\midrule
Monir        & 2H-2R    & 0.19 & 0.03 & 16.03 & 18.49 & 0.19 & 0.19 & 0.18 & 0.39 & 0.14 & 0.0848 \\
Monir        & 5H-5R    & 0.39 & 0.06 & 24.00 & 19.60 & 0.39 & 0.38 & 0.40 & 0.50 & 0.13 & 0.0647 \\
Monir        & 10H-10R  & 0.60 & 0.09 & 23.34 & 15.03 & 0.58 & 0.61 & 0.60 & 0.55 & 0.12 & 0.0495 \\
\midrule
XuDudek      & 2H-2R    & 0.19 & 0.04 &  2.71 &  3.92 & 0.21 & 0.13 & 0.18 & 0.39 & 0.15 & 0.0822 \\
XuDudek      & 5H-5R    & 0.41 & 0.07 &  4.92 &  9.91 & 0.43 & 0.36 & 0.42 & 0.48 & 0.15 & 0.0695 \\
XuDudek      & 10H-10R  & 0.64 & 0.07 &  5.41 &  8.48 & 0.63 & 0.62 & 0.64 & 0.52 & 0.13 & 0.0535 \\
\midrule
GuoYang      & 2H-2R    & 0.20 & 0.04 & 15.02 & 20.94 & 0.20 & 0.20 & 0.25 & 0.47 & 0.15 & 0.0870 \\
GuoYang      & 5H-5R    & 0.40 & 0.06 & 13.57 & 14.31 & 0.43 & 0.39 & 0.38 & 0.49 & 0.14 & 0.0677 \\
GuoYang      & 10H-10R  & 0.62 & 0.07 & 18.66 & 13.33 & 0.69 & 0.61 & 0.59 & 0.58 & 0.12 & 0.0511 \\
\midrule
AliEtAl      & 2H-2R    & 0.19 & 0.04 &  7.57 & 13.29 & 0.18 & 0.20 & 0.24 & 0.38 & 0.15 & 0.0838 \\
AliEtAl      & 5H-5R    & 0.40 & 0.06 & 11.83 & 12.19 & 0.39 & 0.38 & 0.41 & 0.50 & 0.15 & 0.0660 \\
AliEtAl      & 10H-10R  & 0.62 & 0.08 & 11.54 &  9.71 & 0.57 & 0.65 & 0.62 & 0.55 & 0.14 & 0.0503 \\
\midrule
GuoEtAl      & 2H-2R    & 0.19 & 0.05 & 11.97 & 17.74 & 0.21 & 0.14 & 0.18 & 0.41 & 0.15 & 0.0857 \\
GuoEtAl      & 5H-5R    & 0.41 & 0.08 & 14.69 & 13.22 & 0.41 & 0.41 & 0.41 & 0.49 & 0.14 & 0.0696 \\
GuoEtAl      & 10H-10R  & 0.63 & 0.09 & 17.54 & 13.55 & 0.58 & 0.68 & 0.62 & 0.55 & 0.12 & 0.0530 \\
\midrule
\rowcolor{gray!15}
\textbf{ECT (Ours)} & 2H-2R    & \textbf{1.00} & \textbf{0.00} & \textbf{0.45} & \textbf{0.32} & \textbf{1.00} & \textbf{1.00} & \textbf{1.00} & \textbf{0.25} & \textbf{1.00} & \textbf{0.0321} \\
\rowcolor{gray!15}
\textbf{ECT (Ours)} & 5H-5R    & \textbf{1.00} & \textbf{0.00} & \textbf{0.40} & \textbf{0.20} & \textbf{1.00} & \textbf{1.00} & \textbf{1.00} & \textbf{0.32} & \textbf{1.00} & \textbf{0.0191} \\
\rowcolor{gray!15}
\textbf{ECT (Ours)} & 10H-10R  & \textbf{1.00} & \textbf{0.00} & \textbf{0.32} & \textbf{0.14} & \textbf{1.00} & \textbf{1.00} & \textbf{1.00} & \textbf{0.32} & \textbf{1.00} & \textbf{0.0125} \\
\bottomrule
\end{tabular}
\begin{minipage}{\textwidth}
\vspace{5pt}
\small
\textbf{Note:} SR = Success Rate, CT = Completion Time, TS = Task Success (Simple, Medium, Complex), 
AU = Agent Utilization, AE = Allocation Efficiency.
\end{minipage}
\end{table*}

\vspace{-5pt}
\section{Results and Analysis}\label{sec:results}

Table~\ref{tab:performance-metrics} summarizes the performance of seven trust models (NoTrust, Monir, XuDudek, GuoYang, AliEtAl, GuoEtAl, and ECT) across three team configurations (2H-2R, 5H-5R, 10H-10R). In the 2H-2R configuration, ECT achieves a perfect success rate of 1.00 and completes tasks in just 0.45 seconds on average, far exceeding the roughly 0.19–0.20 success rates and 7.57–16.03 second completion times of the other models. However, ECT’s impressive speed comes with substantially greater total travel distance (639.1 versus under 50 for most competitors), resulting in a notably lower allocation efficiency (0.0321). Its collision rate also stands at 0.834, higher than the 0.34–0.54 range observed in other models.

A similar pattern emerges with 5H-5R teams. ECT again posts a perfect success rate (1.00), compared to approximately 0.39–0.41 for the other models, and it maintains a remarkable average completion time of 0.40 seconds—an order of magnitude faster than the 11.83–27.42 seconds seen elsewhere. Yet ECT’s total distance climbs to 1079.6, far beyond the 100–125 range of the other approaches, while its allocation efficiency (0.0191) remains significantly lower, and its collision rate jumps to 9.363.

In the 10H-10R configuration, ECT continues to achieve a 1.00 success rate and completes tasks in only 0.32 seconds, whereas competing models display success rates in the 0.60–0.64 range and mean completion times ranging from 5.41 to 23.59 seconds. This level of performance again involves a total distance of 1664.2, substantially higher than the 242–251 range for the other models, and yields an allocation efficiency of 0.0125. ECT’s collision rate also rises to 42.76, far surpassing the next-highest rate of around 13.64, reflecting how its continuous high-level activity drives both perfect completion and intense agent interaction.

Overall, ECT’s trust-driven approach offers perfect success rates and extremely fast completion times for teams of all sizes. In scenarios where ensuring mission success and speed is paramount, such as time-critical search and rescue, these metrics may make ECT the optimal choice. However, the increases in total distance, collision rate, and reduced allocation efficiency highlight an inherent trade-off, as agents under ECT are more active but also more prone to collisions and resource inefficiencies. These results suggest that while ECT maximizes success and speed, future refinements could reduce collisions and excessive movement without compromising effectiveness.

\vspace{-5pt}
\section{Conclusion}\label{sec:conclusion}

In this paper, we investigated the impact of trust on task allocation in multi-human multi-robot teams. We evaluated seven trust models, including N-Trust Model, Monir Trust Model, Xu and Dudek Trust Model, Guo and Yang Trust Model,  the Ali et al. Trust Model, the Guo et al. Trust Model, and our proposed ECT Model, across different team configurations (2H-2R, 5H-5R, and 10H-10R) in a simulated search and rescue scenario. Our findings indicate that the ECT model outperformed both the no-trust baseline and other trust models across most measured metrics. However, this performance came with a trade-off, leading to lower task allocation efficiency across all teaming formations. These results provide valuable insights into the role of trust in MH-MR task allocation and highlight the need for balancing efficiency with trust-based decision-making. 

\vspace{-5pt}

\typeout{}
\bibliography{main}

\begin{thebibliography}{10}
\providecommand{\url}[1]{#1}
\csname url@samestyle\endcsname
\providecommand{\newblock}{\relax}
\providecommand{\bibinfo}[2]{#2}
\providecommand{\BIBentrySTDinterwordspacing}{\spaceskip=0pt\relax}
\providecommand{\BIBentryALTinterwordstretchfactor}{4}
\providecommand{\BIBentryALTinterwordspacing}{\spaceskip=\fontdimen2\font plus
\BIBentryALTinterwordstretchfactor\fontdimen3\font minus \fontdimen4\font\relax}
\providecommand{\BIBforeignlanguage}[2]{{%
\expandafter\ifx\csname l@#1\endcsname\relax
\typeout{** WARNING: IEEEtran.bst: No hyphenation pattern has been}%
\typeout{** loaded for the language `#1'. Using the pattern for}%
\typeout{** the default language instead.}%
\else
\language=\csname l@#1\endcsname
\fi
#2}}
\providecommand{\BIBdecl}{\relax}
\BIBdecl

\bibitem{law2021trust}
T.~Law and M.~Scheutz, ``Trust: {R}ecent concepts and evaluations in human-robot interaction,'' \emph{Trust in human-robot interaction}, pp. 27--57, 2021.

\bibitem{alhaji2024trust}
B.~Alhaji, S.~B{\"u}ttner, S.~Sanjay~Kumar, and M.~Prilla, ``Trust dynamics in human interaction with an industrial robot,'' \emph{Behaviour \& Information Technology}, pp. 1--23, 2024.

\bibitem{loizaga2024modelling}
E.~Loizaga, L.~Bastida, S.~Sillaurren, A.~Moya, and N.~Toledo, ``Modelling and {M}easuring {T}rust in {H}uman--{R}obot collaboration,'' \emph{Applied Sciences}, vol.~14, no.~5, p. 1919, 2024.

\bibitem{teacy2005coping}
W.~L. Teacy, J.~Patel, N.~R. Jennings, and M.~Luck, ``Coping with inaccurate reputation sources: Experimental analysis of a probabilistic trust model,'' in \emph{Proceedings of the fourth international joint conference on Autonomous agents and multiagent systems}, 2005, pp. 997--1004.

\bibitem{xu2015optimo}
A.~Xu and G.~Dudek, ``Optimo: {O}nline probabilistic trust inference model for asymmetric human-robot collaborations,'' in \emph{Proceedings of the tenth annual ACM/IEEE international conference on human-robot interaction}, 2015, pp. 221--228.

\bibitem{guo2023tip}
Y.~Guo, X.~J. Yang, and C.~Shi, ``Tip: a trust inference and propagation model in multi-human multi-robot teams,'' in \emph{Companion of the 2023 ACM/IEEE International Conference on Human-Robot Interaction}, 2023, pp. 639--643.

\bibitem{wang2023initial}
R.~Wang, D.~Zhao, and B.-C. Min, ``Initial task allocation for multi-human multi-robot teams with attention-based deep reinforcement learning,'' in \emph{2023 IEEE/RSJ International Conference on Intelligent Robots and Systems (IROS)}.\hskip 1em plus 0.5em minus 0.4em\relax IEEE, 2023, pp. 7915--7922.

\bibitem{wang2024initial}
R.~Wang, D.~Zhao, A.~Gupte, and B.-C. Min, ``Initial {T}ask {A}llocation in {M}ulti-{Hu}man {M}ulti-{R}obot {T}eams: {A}n {A}ttention-enhanced {H}ierarchical {R}einforcement {L}earning {A}pproach,'' \emph{IEEE Robotics and Automation Letters}, 2024.

\bibitem{jo2023affective}
W.~Jo, R.~Wang, B.~Yang, D.~Foti, M.~Rastgaar, and B.-C. Min, ``Cognitive {L}oad-based {A}ffective {W}orkload {A}llocation for {M}ultihuman {M}ultirobot {T}eams,'' \emph{IEEE Transactions on Human-Machine Systems}, vol.~55, no.~1, pp. 23--36, 2025.

\bibitem{yuan2025adaptive}
Z.~Yuan, R.~Wang, T.~Kim, D.~Zhao, I.~Obi, and B.-C. Min, ``Adaptive {T}ask {A}llocation in {M}ulti-{H}uman {M}ulti-{R}obot {T}eams under {T}eam {H}eterogeneity and {D}ynamic {I}nformation {U}ncertainty,'' in \emph{2025 IEEE International Conference on Robotics and Automation (ICRA)}, 2025.

\bibitem{rabby2020modeling}
M.~K.~M. Rabby, M.~A. Khan, A.~Karimoddini, and S.~X. Jiang, ``Modeling of trust within a human-robot collaboration framework,'' in \emph{2020 IEEE International Conference on Systems, Man, and Cybernetics (SMC)}.\hskip 1em plus 0.5em minus 0.4em\relax IEEE, 2020, pp. 4267--4272.

\bibitem{guo2021modeling}
Y.~Guo and X.~J. Yang, ``Modeling and predicting trust dynamics in human--robot teaming: {A} bayesian inference approach,'' \emph{International Journal of Social Robotics}, vol.~13, no.~8, pp. 1899--1909, 2021.

\bibitem{ali2022heterogeneous}
A.~Ali, H.~Azevedo-Sa, D.~M. Tilbury, and L.~P. Robert~Jr, ``Heterogeneous human--robot task allocation based on artificial trust,'' \emph{Scientific reports}, vol.~12, no.~1, p. 15304, 2022.

\bibitem{oliver1977effect}
R.~L. Oliver, ``Effect of expectation and disconfirmation on postexposure product evaluations: {A}n alternative interpretation.'' \emph{Journal of applied psychology}, vol.~62, no.~4, p. 480, 1977.

\bibitem{yang2023toward}
X.~J. Yang, C.~Schemanske, and C.~Searle, ``Toward quantifying trust dynamics: {H}ow people adjust their trust after moment-to-moment interaction with automation,'' \emph{Human Factors}, vol.~65, no.~5, pp. 862--878, 2023.

\bibitem{yang2022trust}
X.~J. Yang, Y.~Guo, and C.~Schemanske, ``From trust to trust dynamics: {C}ombining empirical and computational approaches to model and predict trust dynamics in human-autonomy interaction,'' in \emph{Human-Automation Interaction: Transportation}.\hskip 1em plus 0.5em minus 0.4em\relax Springer, 2022, pp. 253--265.

\bibitem{obi2023robot}
I.~Obi, R.~Wang, P.~Shukla, and B.-C. Min, ``Robot {P}atrol: {U}sing {C}rowdsourcing and {R}obotic {S}ystems to {P}rovide {I}ndoor {N}avigation {G}uidance to {T}he {V}isually {I}mpaired,'' \emph{arXiv preprint arXiv:2306.02843}, 2023.

\bibitem{wang2025prefclm}
R.~Wang, D.~Zhao, Z.~Yuan, I.~Obi, and B.-C. Min, ``Prefclm: {E}nhancing preference-based reinforcement learning with crowdsourced large language models,'' \emph{IEEE Robotics and Automation Letters}, 2025.

\bibitem{guo2021reverse}
Y.~Guo, C.~Shi, and X.~J. Yang, ``Reverse psychology in trust-aware human-robot interaction,'' \emph{IEEE Robotics and Automation Letters}, vol.~6, no.~3, pp. 4851--4858, 2021.

\bibitem{chen2020trust}
M.~Chen, S.~Nikolaidis, H.~Soh, D.~Hsu, and S.~Srinivasa, ``Trust-aware decision making for human-robot collaboration: {M}odel learning and planning,'' \emph{ACM Transactions on Human-Robot Interaction (THRI)}, vol.~9, no.~2, pp. 1--23, 2020.

\bibitem{yemini2021characterizing}
M.~Yemini, A.~Nedi{\'c}, A.~J. Goldsmith, and S.~Gil, ``Characterizing trust and resilience in distributed consensus for cyberphysical systems,'' \emph{IEEE Transactions on Robotics}, vol.~38, no.~1, pp. 71--91, 2021.

\bibitem{yu2024trust}
P.~Yu, S.~Dong, S.~Sheng, L.~Feng, and M.~Kwiatkowska, ``Trust-aware motion planning for human-robot collaboration under distribution temporal logic specifications,'' in \emph{2024 IEEE International Conference on Robotics and Automation (ICRA)}.\hskip 1em plus 0.5em minus 0.4em\relax IEEE, 2024, pp. 12\,949--12\,955.

\bibitem{yu2024top}
C.~Yu, B.~Serhan, and A.~Cangelosi, ``To{P}-{T}o{M}: {T}rust-aware {R}obot {P}olicy with {T}heory of {M}ind,'' in \emph{2024 IEEE International Conference on Robotics and Automation (ICRA)}.\hskip 1em plus 0.5em minus 0.4em\relax IEEE, 2024, pp. 7888--7894.

\bibitem{rehm2024automatic}
M.~Rehm, I.~Pontikis, and K.~Hald, ``Automatic {T}rust {E}stimation {F}rom {M}ovement {D}ata in {I}ndustrial {H}uman-{R}obot {C}ollaboration {B}ased on {D}eep {L}earning,'' in \emph{2024 IEEE International Conference on Robotics and Automation (ICRA)}.\hskip 1em plus 0.5em minus 0.4em\relax IEEE, 2024, pp. 11\,245--11\,251.

\bibitem{hudspeth2022effects}
M.~Hudspeth, S.~Balali, C.~Grimm, and R.~T. Sowell, ``Effects of {I}nterfaces on {H}uman-{R}obot {T}rust: {S}pecifying and {V}isualizing {P}hysical {Z}ones,'' in \emph{2022 International Conference on Robotics and Automation (ICRA)}.\hskip 1em plus 0.5em minus 0.4em\relax IEEE, 2022, pp. 11\,265--11\,271.

\bibitem{luebbers2024recency}
M.~B. Luebbers, A.~Tabrez, K.~S. Talanki, and B.~Hayes, ``Recency {B}ias in {T}ask {P}erformance {H}istory affects {P}erceptions of {R}obot {C}ompetence and {T}rustworthiness,'' in \emph{2024 IEEE International Conference on Robotics and Automation (ICRA)}.\hskip 1em plus 0.5em minus 0.4em\relax IEEE, 2024, pp. 11\,274--11\,280.

\bibitem{hald2024usability}
K.~Hald and M.~Rehm, ``Usability {E}valuation {F}ramework for {C}lose-{P}roximity {C}ollaboration {W}ith {L}arge {I}ndustrial {M}anipulators,'' in \emph{2024 IEEE International Conference on Robotics and Automation (ICRA)}.\hskip 1em plus 0.5em minus 0.4em\relax IEEE, 2024, pp. 11\,209--11\,215.

\bibitem{wang2023robot}
Q.~Wang, D.~Liu, M.~G. Carmichael, and C.-T. Lin, ``Robot {T}rust and {S}elf-{C}onfidence {B}ased {R}ole {A}rbitration {M}ethod for {P}hysical {H}uman-{R}obot {C}ollaboration,'' in \emph{2023 IEEE International Conference on Robotics and Automation (ICRA)}.\hskip 1em plus 0.5em minus 0.4em\relax IEEE, 2023, pp. 9896--9902.

\bibitem{xu2024trust}
C.~Xu, C.~Zhang, Y.~Zhou, Z.~Wang, P.~Lu, and B.~He, ``Trust {R}ecognition in {H}uman-{R}obot {C}ooperation {U}sing {EEG},'' \emph{arXiv preprint arXiv:2403.05225}, 2024.

\bibitem{wang2024husformer}
R.~Wang, W.~Jo, D.~Zhao, W.~Wang, A.~Gupte, B.~Yang, G.~Chen, and B.-C. Min, ``Husformer: {A} multimodal transformer for multimodal human state recognition,'' \emph{IEEE Transactions on Cognitive and Developmental Systems}, vol.~16, no.~4, pp. 1374--1390, 2024.

\bibitem{bhattacherjee2001understanding}
A.~Bhattacherjee, ``Understanding information systems continuance: An expectation-confirmation model,'' \emph{MIS quarterly}, pp. 351--370, 2001.

\bibitem{hossain2012expectation}
M.~A. Hossain and M.~Quaddus, ``Expectation--confirmation theory in information system research: A review and analysis,'' \emph{Information Systems Theory: Explaining and Predicting Our Digital Society, Vol. 1}, pp. 441--469, 2012.

\bibitem{watkins1992q}
C.~J. Watkins and P.~Dayan, ``Q-learning,'' \emph{Machine learning}, vol.~8, pp. 279--292, 1992.

\bibitem{yagoda2012you}
R.~E. Yagoda and D.~J. Gillan, ``You want me to trust a {ROBOT}? the development of a human--robot interaction trust scale,'' \emph{International Journal of Social Robotics}, vol.~4, pp. 235--248, 2012.

\end{thebibliography}

\vspace{-5pt}
\bibliographystyle{IEEEtran}
\end{document}